\documentclass{bmvc2k}

\title{Sampling Based On Natural Image Statistics Improves Local Surrogate Explainers}

\addauthor{Ricardo Kleinlein}{ricardo.kleinlein@upm.es}{1}
\addauthor{Alexander Hepburn}{ah13558@bristol.ac.uk}{2}
\addauthor{Raúl Santos-Rodríguez}{enrsr@bristol.ac.uk}{2}
\addauthor{Fernando Fernández-Martínez}{fernando.fernandezm@upm.es}{1}

\addinstitution{
Information Processing and Telecommunications Center\\
E.T.S.I. de Telecomunicación,Universidad Politécnica de Madrid \\
Madrid, Spain
}
\addinstitution{
Department of Engineering Mathematics \\
University of Bristol \\
Bristol, United Kingdom
}

\runninghead{Kleinlein et al.}{Sampling Based On Natural Image Statistics}

\DeclareMathOperator*{\argmin}{arg\,min}

\begin{document}

\maketitle

\begin{abstract}
Many problems in computer vision have recently been tackled using models whose predictions cannot be easily interpreted, most commonly deep neural networks.
Surrogate explainers are a popular post-hoc interpretability method to further understand how a model arrives at a particular prediction. 
By training a simple, more interpretable model to locally approximate the decision boundary of a non-interpretable system, we can estimate the relative importance of the input features on the prediction. Focusing on images, surrogate explainers, e.g., LIME, generate a local neighbourhood around a query image by sampling in an interpretable domain.
However, these interpretable domains have traditionally been derived exclusively from the intrinsic features of the query image, not taking into consideration the manifold of the data the non-interpretable model has been exposed to in training (or more generally, the manifold of real images). This leads to suboptimal surrogates trained on potentially low probability images.
We address this limitation by aligning the local neighbourhood on which the surrogate is trained with the original training data distribution, even when this distribution is not accessible. We propose two approaches to do so, namely (1) altering the method for sampling the local neighbourhood and (2) using perceptual metrics to convey some of the properties of the distribution of natural images.

\end{abstract}


\section{Introduction}
\label{sec:intro}

\begin{figure}[htb]
\centering
\begin{tabular}{cc}
    \textbf{Undistorted source image} & \textbf{Distorted (blurred) version} \\
    \includegraphics[width=0.4\textwidth]{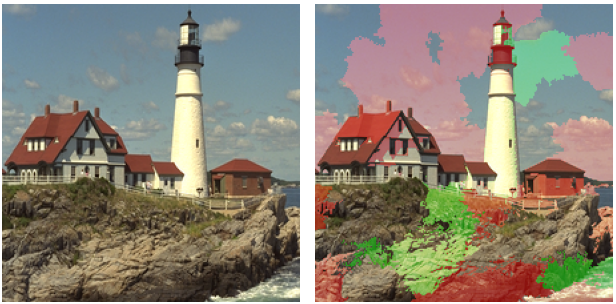} &
    \includegraphics[width=0.4\textwidth]{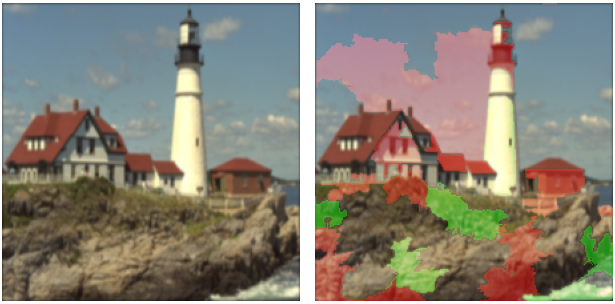} \\
    \multicolumn{2}{c}{(a) Default LIME: cosine distance \& mean color occlusion} \\
    \includegraphics[width=0.4\textwidth]{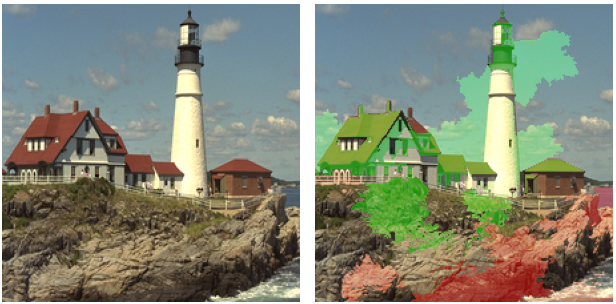} &
    \includegraphics[width=0.4\textwidth]{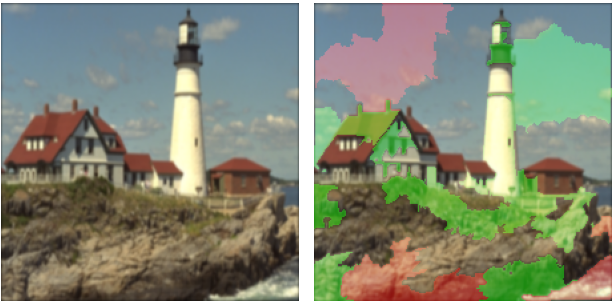} \\
    \multicolumn{2}{c}{(b) Proposed explanations using MS-SSIM \& aligned sampling (blurring)} \\
    \\
\end{tabular}
\label{fig:overview}
\caption{Explanations provided for both clear and distorted pictures labeled as  "lighthouse". To align the distance metric and the sampling method to the statistics of real images not only makes explanations more similar to what humans would estimate important to understand the prediction but also enhances the robustness of the explanation.} 
\end{figure}

Deep neural networks are at the forefront of both computer vision research and related industrial applications~\cite{deng2009imagenet, He_2016_CVPR}.
Practitioners in the field often use these large uninterpretable models due to their capacity to make accurate predictions even at the cost of not being able to understand the rationale behind a prediction.
Therefore, if a particular prediction does not match our expectations, we lack the resources to assess the reasoning behind it, hence undermining the confidence of the users in the whole system~\cite{Miller2019ExplanationIA}. 

\begin{figure}[htb]
    \centering
    \includegraphics[width=0.84\textwidth]{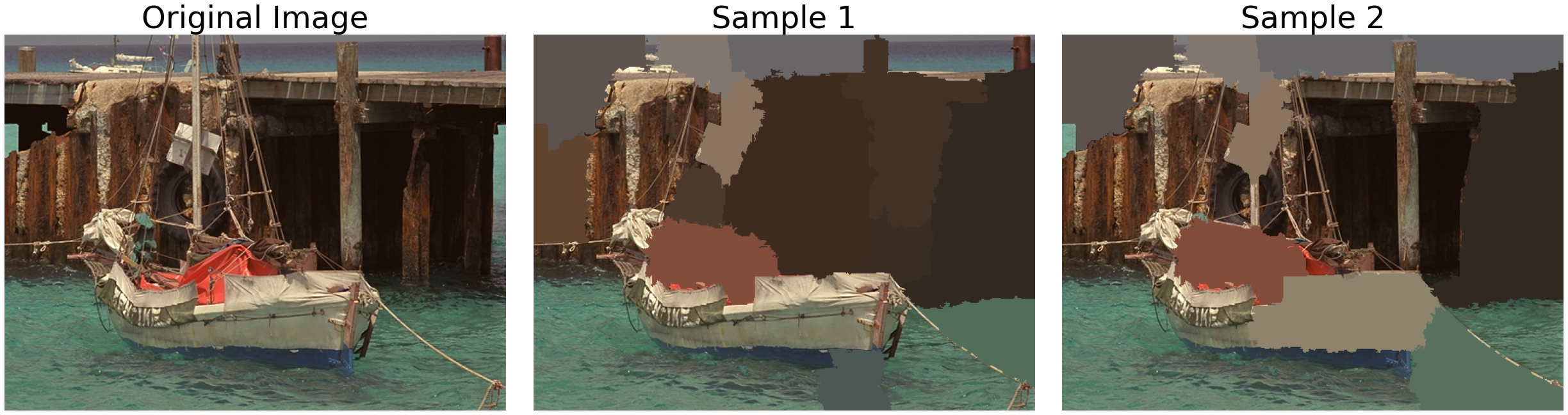}
    \caption{The original image (left) and examples of sampled data using the LIME approach, with mean colour occlusion.}
    \label{fig:lime_sampling_eg}
\end{figure}

Even though the exact definition is still a matter of debate~\cite{Lipton2018}, "explaining a prediction" generally means presenting visual cues that allow users to build an intuition on the input features that drive the decision-making process of a model, be that prediction either wrong or right. 
Post-hoc interpretability techniques range from, counterfactuals -- finding the closest data point of the opposite class~\cite{wachter2017counterfactual} -- to  permuting values of a feature to check the effect this has on a classification prediction~\cite{friedman2001greedy}. Alternatively, surrogate explainers involve locally approximating the decision boundary around a query point, using a simpler interpretable model~\cite{NIPS1995Craven,Ribeiro2016}.


In this work, we focus on surrogate explainers for image tasks. In this context, images are usually represented as a collection of super-pixels that encapsulate adjacent pixel regions with similar visual properties. In order to build an interpretable domain and train the corresponding surrogate, a neighbourhood is sampled around the query image. As seen in Fig.~\ref{fig:lime_sampling_eg}, because of the generation techniques, the samples are far from what we can consider real images. We argue that is process is sub-optimal and leads to inconsistent explanations. We suggest that the more this neighbourhood resembles the distribution that the global model was trained using (or at least the distribution of natural images), the more robust the explanations will be. We propose two ways of achieving this. First, using a different sampling method to better exploit the training distribution (if available) or using perceptual metrics that have been shown to encode fundamental properties of the distribution of natural images. In general, when considering images, access to the actual distribution is intractable but it has been shown that perceptual metrics are correlated with the probability of natural images~\cite{hepburn2022on}, acting as proxy to a space that is more aligned with the distribution of natural images than the Euclidean space. 

\section{Surrogate Explainers}
\label{sec:related-work}

Although it is a custom practise to evaluate the adequacy of a machine learning model according to a fixed set of metrics like accuracy rates for classification, or mean squared error for regression tasks, these are not sufficient to fully characterise its behaviour.
Because of this limitation, a complete understanding of an automatic system should also include the ability to explain its predictions~\cite{Lundberg2019}.
Arguably, the most popular approach to provide explanations for black-box models are surrogate-based techniques.
These are post-hoc local approximations to the decision boundary of a black-box system around a query point $x$, learned by a simple model that is, in many cases,  linear.
We define an explanation as $\text{exp}_{x} = \argmin_{g \in \mathcal{G}}\mathcal{L}(f, g, \pi_x) + \Omega(g)$, where $\mathcal{L}(f, g, \pi_x)$ is the fit of the surrogate model $g$ from the family of surrogates $G$, $f$ denotes the black-box model and $\pi_x$ is the neighbourhood sampled around a query data point $x \in \mathcal{X}$.
$\Omega(g)$ is a penalty on the complexity of model $g$.
If $\Omega(g)$ is the L2 norm of the weights $g$, the surrogate model becomes ridge regression.
This formulation conveys the idea that our surrogate $g$ should find the best fit to the local boundary decision of the black-box $f$ given the restrictions in complexity expressed by $\Omega(g)$, only around the neighbourhood $\pi_x$. Surrogate explainers can be broken up into three interoperable modules; an interpretable data representation, a data sampling procedure and the explanation generation~\cite{sokol2019blimey}. These modules are required to be carefully chosen by the practitioner in order to address the problem at hand.

\paragraph{Interpretable data representation} Samples in the original domain $\mathcal{X}$ are transformed into a human interpretable representation, $\mathcal{Z}$. In the case of natural images, superpixels offer an interpretable domain in which an image can be represented as a binary vector that encodes whether a specific region has been occluded (removed) or altered in any way.

\paragraph{Sampling} Data sampling refers to the generation of the neighbourhood $\pi_x$. For images, this usually involves sampling binary vectors in the interpretable data representation defined above (see Fig. \ref{fig:overview}).

\paragraph{Explanation generation} The final stage is training the surrogate model in order to replicate the behaviour of the black-box model. The surrogate model aims to learning the mapping between instances in $\pi_x$ sampled from the interpretable domain $z \in \mathcal{Z}$ and the probabilities estimated for a given class by the black-box model, $f(z)$.
Consequently, we define a locally weighted square loss, $\mathcal{L}(f, g, \pi_x) = \sum_{z, z' \in \mathcal{Z}} \pi_x (x, z)(f(z) - g(z'))^2$.

\section{Surrogates and Natural Image Statistics}
\label{sec:statistics}

Even though it is clear that the choice of $\pi_x$ is critical when producing reliable explanations, to our knowledge, barely any studies have investigated this topic~\cite{Hepburn2021,poyiadzi2020}. We believe that in order to faithfully mimic the vicinity around a query $x$ to be explained, $\pi_x$ should incorporate information about the statistics of the distribution the black-box model was trained on. A possible way to do so is to change the sampling process in the interpretable domain $\mathcal{Z}$.

\begin{figure*}
    \centering
    \begin{tabular}{ccc}
         \includegraphics[scale=0.3]{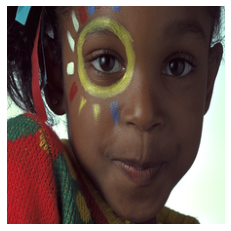} &
         \includegraphics[scale=0.3]{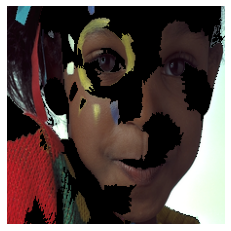} &
         \includegraphics[scale=0.3]{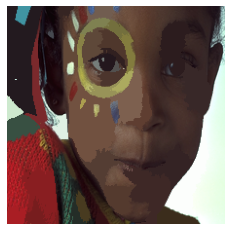} \\
         (a) Source image & (b) LIME  & (c) LIME \\
         & (black occlusion) & (mean color occlusion) \\
          \includegraphics[scale=0.3]{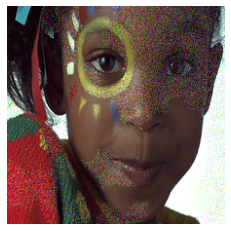} &
         \includegraphics[scale=0.3]{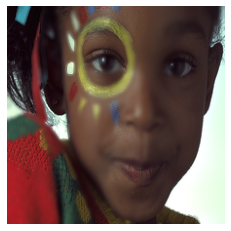} &
         \includegraphics[scale=0.3]{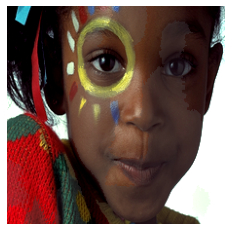} \\
         (d) Additive  & (e) Gaussian & (f) Contrast \\
         gaussian noise & blurring & change \\

    \end{tabular}
    \caption{Images sampled from the same binary representation of superpixels. Contrarily to LIME-like patching methods, we sample from distributions closer to that of real images.}
    \label{fig:sampling-images}
\end{figure*}

\subsection{Sampling according to the statistics of natural images}

Sampling is usually performed in the interpretable domain $\mathcal{Z}$.
This representation usually has a one-to-one correspondence with the original, non-interpretable domain $\mathcal{X}$ since, in order to train a surrogate, the output of the black-box model $f(\pi_x)$ is required. For tabular data, LIME suggests using a one-to-many mapping and performs inverse sampling, where a Gaussian is fit to the data in one area of $\mathcal{X}$, corresponding to one integer value in the interpretable domain $\mathcal{Z}$. However, this leads to increased randomness and variability between the surrogates produced~\cite{sokol2019blimey, fen2019should}.

Applying LIME to images, the domain of superpixels can be sampled as binary vectors $z'$ from a discrete uniform distribution that results in images $z$ with superpixels ablated according to the binary feature in the sampled vectors.
Usually, this binary representation encodes either whether to set all pixel values within an ablated superpixel to 0 (zero patching), or to set these pixels to their mean value within the superpixel (mean colour occlusion).
We propose to use more realistic sampling methods in order to capture the statistics of real-world images. The core idea being if I remove the information in this superpixel, what effect will it have on the prediction.

Given direct access to the actual distribution of real-world images, we could replace the ablated superpixel with samples drawn from such distribution. As this is usually intractable, we propose to approximate the effect by transforming the pixels according to distortions often found in natural images.
As can be seen from Figure~\ref{fig:sampling-images}, sampling from the proposed distributions leads to more realistic images, with the neighbourhood $\pi_x$ being closer to the actual distribution used to train the the global model.
Alternatively, perceptual metrics that naturally encapsulate properties about the distribution of real images can be explored.

\subsection{Perceptual metrics}
\label{subsec: perceptual}

Another fundamental aspect when sampling a neighbourhood $\pi_x$ is to measure the distance between the generated samples and the query $x$. A neighbourhood is defined by $
    \pi_x(x, z) = \text{exp}\left(-\frac{D(x, z)^2}{\sigma^2}\right)
$
where $D$ is a distance between samples $z$ and $x$ and $\sigma$ is the width of the exponential kernel. 
The distance used in the original implementation of LIME is the cosine similarity between a vector of all ones representing the superpixels of the original image $x'$ and the binary mask of superpixels in sample $z'$. However, as image explanations come in the form of visual cues, the explanation we produce must ultimately resemble the way we humans interpret and think about visual information. As a consequence, the cosine distance may not be the best choice in all scenarios. This is particularly important for natural images, where properties related to the psychophysics of human perception take a predominant role in explaining a prediction. Perceptual metrics attempt to recreate human psychophysical results when observing a reference and distorted image. Models based on the human visual system have been proved to be effective at this task~\cite{Wang2003, laparra2016perceptual, martinez2018derivatives}.
One such metric is structural similarity and its multi-scale variant, the \textit{multi-scale structural similarity} index (MS-SSIM)~\cite{Wang2003}. MS-SSIM is based on the principle that the perceived structural similarity will be preserved despite the distortion.
It has been shown that this distance faithfully reflects the perceptual similarity between two images as perceived by the human visual system\cite{hepburn2022on}.
In our empirical analysis on natural images we will use MS-SSIM as $D(x, z)$ as an alternative to the cosine distance in order to implicitly alter the distribution of sampled images, making it closer to that of the training data of the black-box model.


\section{Experiments}
\label{sec:experimentsimages}

The visual domain poses a particularly challenging environment since sampling from the actual data distribution of all the  real images is far from trivial.
Typically we do not have access to that distribution and hence we resort to alternative ways of reconstructing the vicinity of an image.
While LIME assumes that the neighbourhood of an image can be approximated by sampling a subset of patched versions of that image, this is generally not true.
In this Section, we first illustrate how to improve upon this restriction by sampling from the real distribution in the context of a toy problem in 2D.
Then, we proceed to show that similar results can be achieved in the visual domain.
However, in both, as we do not know the distribution of all possible real images, rather than sampling from it, we need to approximate it.
Furthermore, we explore the use of perceptual metrics as an alternative proxy to enforce a similar behaviour.
In all cases, a ridge regressor works as our surrogate model.


\subsection{Synthetic example}
\label{sec:experiments2d}

First, we test how altering the sampling can affect the resulting surrogate explainer in a simple 2D case. We use a non-parametric uniformisation transform in order to have a gradual scale between sampling independent of the distribution, a bounded uniform distribution around our query point, and according to the original data distribution. We explore the impact this has on the resulting surrogate in a dimensionality which we can visualise before presenting experiments using real images. We use a two moons dataset, with additive Gaussian noise with a standard deviation of 0.35 to ensure the classes are linearly inseparable. A random forest is trained on samples taken from this distribution.

\paragraph{Approximating the data distribution}
In order to achieve a scale between sampling according to the distribution or independently of it, we use a quantile transformation. An estimate of the cumulative distribution function is used to map the values to a uniform distribution. The number of quantiles, or the number of points the CDF is estimated using, dictates the degree of uniformisation. A high number of quantiles leads to a more accurate estimation of the CDF and a more uniform transformation. To sample our data for the surrogate explainer, we initially sample uniformly within certain bounds around the query point in order to enforce locality. We then progressively use the inverse transform of the quantile uniformisation, with an increasing number of quantiles. The result is data that is transformed to be more like the original distribution the higher the number of quantiles used. Surrogate explainers are then obtained using this sampled data, and we observe how different these surrogates are compared to a surrogate trained on data from the true distribution. 

\paragraph{Evaluation}
The Wasserstein distance is measured between the sampled data and samples from the true distribution within the same bounds. The $\ell_2$ distance between parameters of the ridge regression are used to track the distance between the obtained explainer, and the surrogate explainer trained on the true distribution.

\paragraph{Experiments}
We first train the quantile uniformisation transforms, with quantiles in the range $[2, 100]$. 50 query points $x$ are chosen from a test set of the two moons distribution, and we sample uniformly around them in both feature directions in the range $[x-\sigma, x+\sigma]$, with $\sigma=0.2$ to ensure that the samples, and the resulting explanation, are local. The inverse uniformisation is then applied (going from a uniform distribution to an approximation of the two moons distribution) and a surrogate is trained on this data.

\begin{figure}
    \centering
    \includegraphics[width=\textwidth]{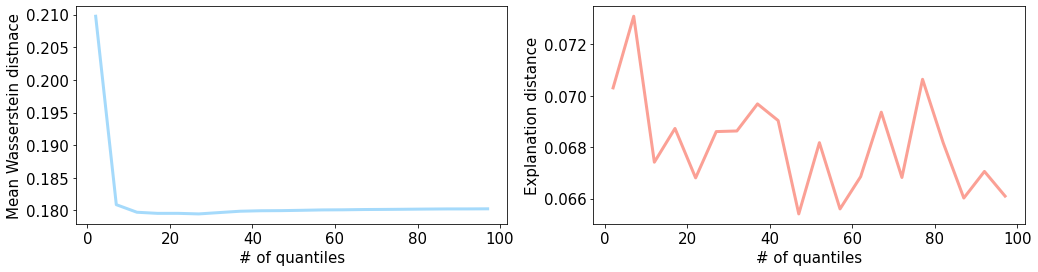}
    \caption{Mean Wasserstein distance and explanation distance as a function of the number of quantiles used to transform the synthetic data.}
    \label{fig:2d_average_results}
\end{figure}

\begin{figure}[ht]
    \centering
    \includegraphics[width=\textwidth]{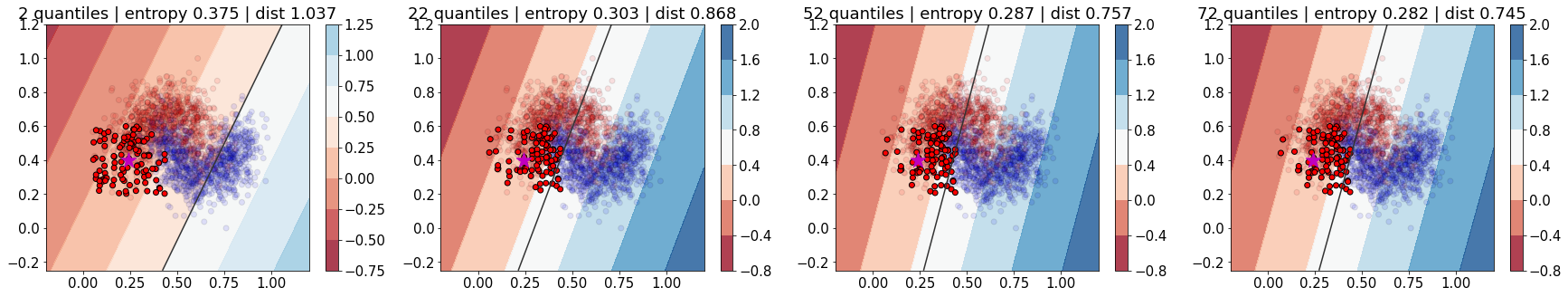}
    \caption{Surrogates trained on the two moons dataset, where the sampled data is transformed to be increasingly similar to the original distribution. Top left resembles uniform sampled data, while bottom right shows sampled data following the two moons distribution.}
    \label{fig:2d_twomoons}
\end{figure}

\paragraph{Results}
Fig.~\ref{fig:2d_twomoons} is an example of the generated surrogates for a point close to the global model boundary and towards the edge of the distribution. Fig.~\ref{fig:2d_average_results} shows the average across 50 query points. 
As we increase the number of quantiles used in the transformation, the Wasserstein distance decreases. This means our sampled data is more similar to the original distribution. The distance between the resulting surrogate and one trained on the true distribution also decreases, meaning that when we sample data increasingly more similar to the the true distribution, the resulting surrogate explainer is also more similar. This may seem somewhat trivial, but it is important to note that this concept can be applied to more complicated distributions, where we cannot sample data easily and then transform according to the distribution.

\subsection{Natural images}

\paragraph{Dataset}
TID2008 is a dataset composed of 25 natural images depicting different objects, landscapes and people, as well as some artificially-generated scenes~\cite{Ponomarenko2009}.
It was originally devised to evaluate visual 
quality assessment metrics in images affected by 17 types of distortion (like adding salt\&pepper noise or having compression errors) at 4 different degrees of intensity.
Due to the computation expenses with generating an explanation, we restrict our experiments to clear samples and their counterparts containing either additive Gaussian noise, Gaussian blurring or a change in contrast.
Hence, we create an explanation for each of the 100 pictures available, for each type of distortion considered.
Images are downsampled to 224 x 224 pixels in all our experiments.

\paragraph{Approximating the data distribution}
Superpixels remain the interpretable data representation in our experiments, yet we extend on the meaning of the binary mask of the original LIME.
In order to mimic the data the black-box model is trained from, instead of setting pixels within a superpixel to their mean value, our binary vectors represent whether pixels are altered according to one of the distortions considered in our dataset (Figure~\ref{fig:sampling-images}).
We either apply additive Gaussian noise with increasing mean intensity values $0.01$, $0.05$ and $0.1$, Gaussian blurring with kernel sizes $3x3$, $5x5$ and $11x11$ or a change in the contrast of the image.
In this case, a change in contrast of $0.5$ halves the original contrast level of an image, whereas a value of $1$ leaves it intact.
Considering the nature of the images to explain and modifying the sampling accordingly, we are leveraging the resources available to make the training data of the surrogate more realistic, as well as more similar to that of the black-box.

\paragraph{Black-box model}
Transformer architectures, although initially introduced to tackle natural language processing tasks~\cite{Vaswani2017AttentionIA, Devlin2019BERTPO}, are also becoming a popular choice in the computer vision domain~\cite{Radford2021LearningTV}.
Due to their adoption by the community, we choose a Vision Transformer (ViT) model trained for image classification as our black-box predictive system~\cite{Dosovitskiy2021AnII}.
The specific implementation we use is a version pretrained on the Imagenet-21K dataset~\cite{Russakovsky2015}.

\paragraph{Distance between explanations}

The evaluation of the quality and reliability of computational explanations of black-box models is a complex problem~\cite{Poyiadzi2021OnTO, Sokol2020}.
To judge the similarity between explanations, we follow the procedure introduced in~\cite{Hepburn2021}, where the authors define the empirical distance between two explanations $\mathbf{E_k}$ and $\mathbf{\tilde{E}_k}$ as
    $D_{exp} = \frac{1}{K} \sum^K_{k=0}| \mathbf{E}_k - \mathbf{\tilde{E}_k} | ^2_F$, 
where the sum is performed over $K$ explained classes.
In our experiments we generate surrogates only for the most likely class predicted by the classification model, thus $K=1$.
$\mathbf{E_k}$ (and analogously, $\mathbf{\tilde{E}_k}$) is a matrix in which each $E_k(i, j)$ denotes the importance value in the explanation of the superpixel the pixel in position $(i, j)$ belongs to for class $k$.

\paragraph{Experiments}
Surrogates for pairs of reference and distorted images are trained using the same combination of sampling procedure and distance metric, to ensure coherence in the explanations generated.
These are then projected as pixel relevance maps and the distance between them is computed using the definition above.
We repeat this procedure for all reference images and average over each type of distortion considered.

\paragraph{Results}

\begin{figure}[t]
    \centering
    \begin{tabular}{c}
    \includegraphics[width=0.8\textwidth]{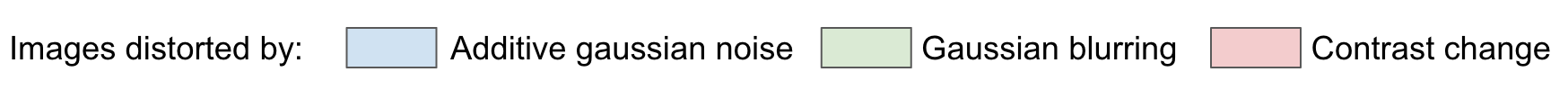} \\
        \includegraphics[width=0.8\textwidth]{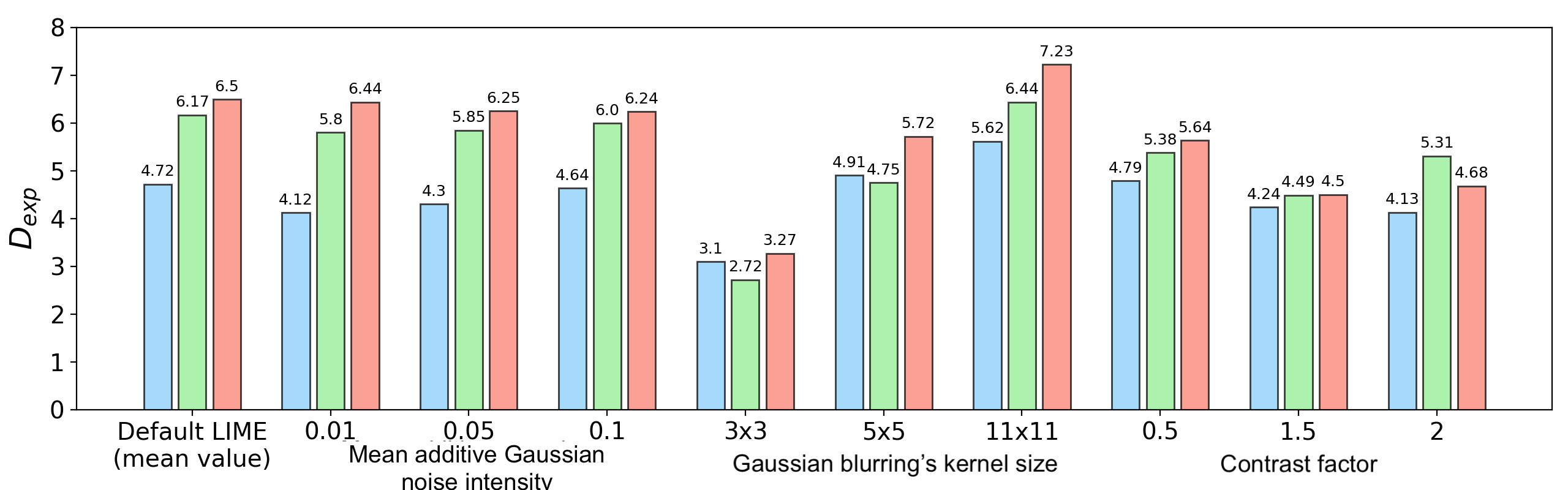} \\
        (a) Using a cosine distance metric. \\
        \includegraphics[width=0.8\textwidth]{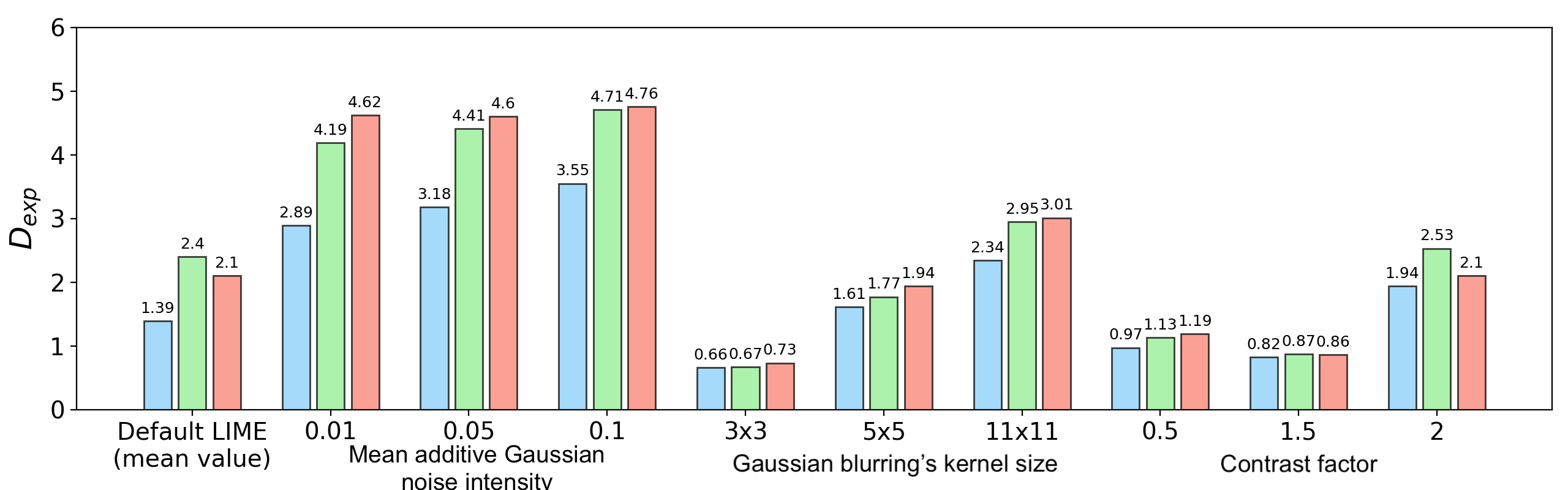} \\
        (b) Using MS-SSIM  as distance metric. \\
    \end{tabular}
    \caption{Average absolute distance computed for each pair of clear-distorted images depending on the sampling method, the distortion type of the image and the distance metric used. The $x$-axis indicates the transform applied to superpixels in the sampling process.}
    \label{fig:raw-image-results}
\end{figure}

\begin{figure}
    \centering
    \begin{tabular}{cc}
     \multicolumn{2}{c}{
        \includegraphics[width=0.8\textwidth]{images/relative-legen2.png}} \\
        \includegraphics[width=0.35\textwidth]{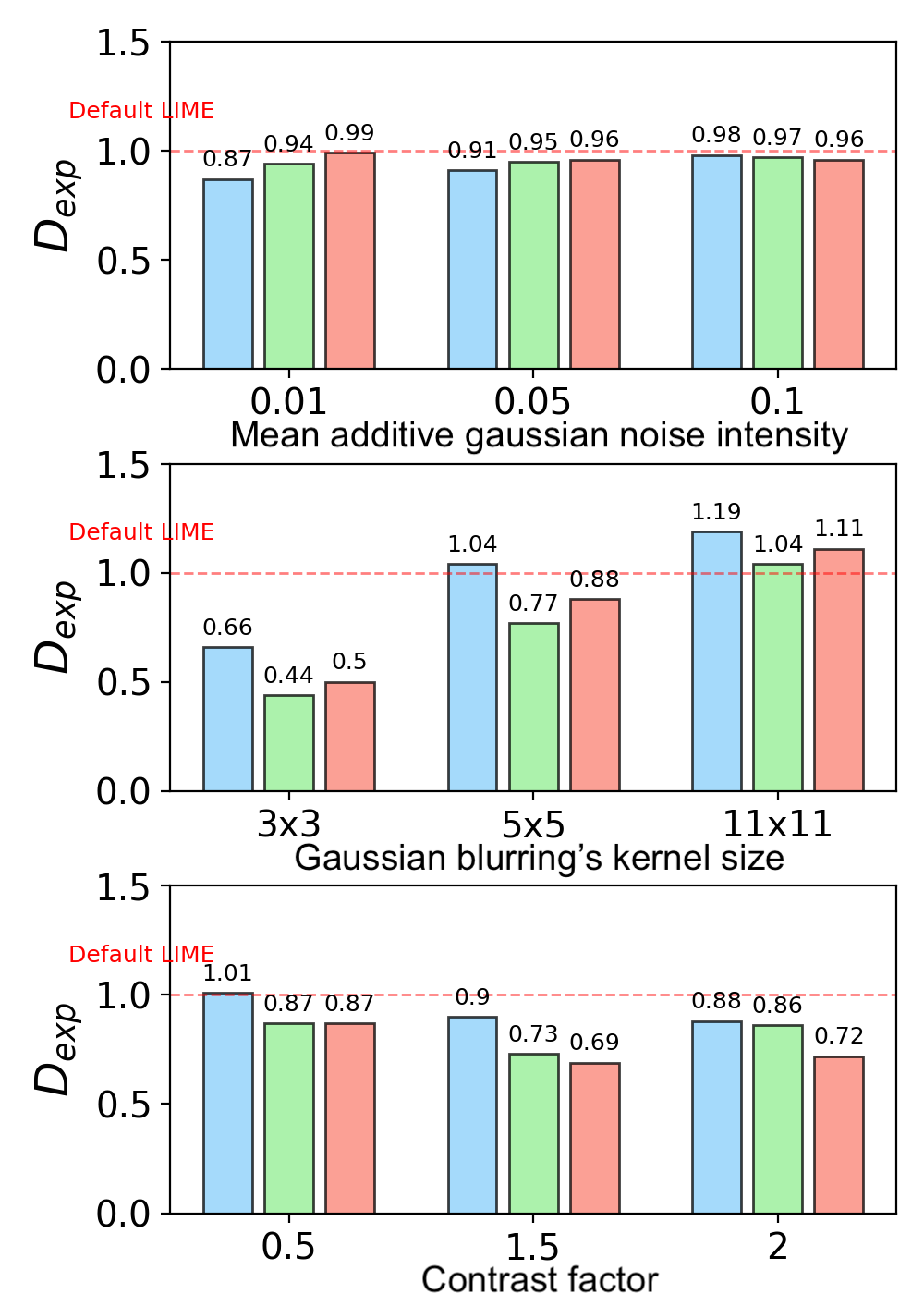} &
        \includegraphics[width=0.35\textwidth]{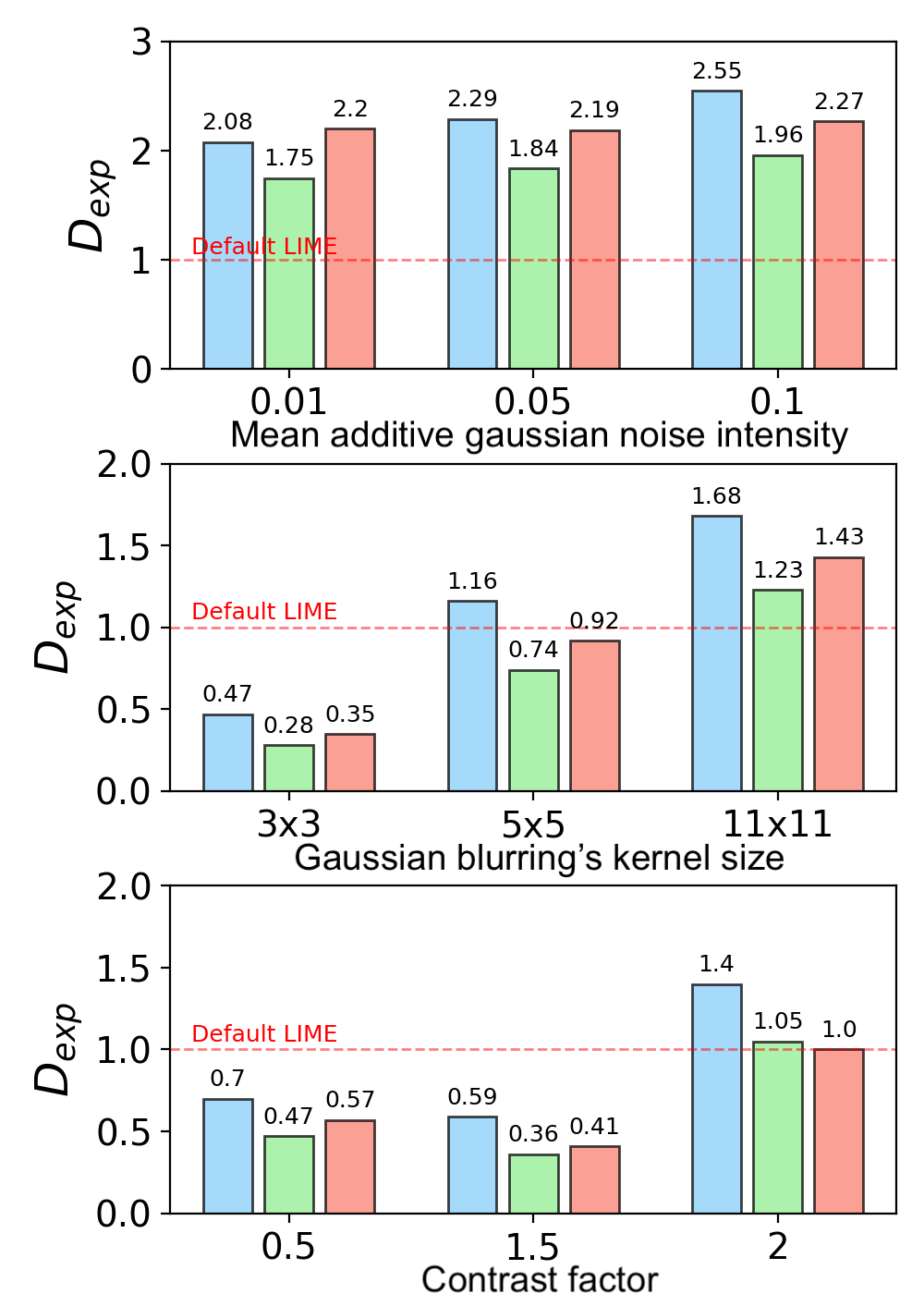} \\
        (a) Cosine distance & (b) MS-SSIM \\
    \end{tabular}
    \caption{Average distance between explanations normalised to the mean distance computed using the default LIME configuration. It is important to align some sampling methods to the nature of the images explained in the absence of access to the true distribution.}
    \label{fig:image-results}
\end{figure}

First of all, it can be seen from Figure~\ref{fig:raw-image-results} that by using a perceptual distance to measure the proximity of the local neighbours, we can partially alleviate the lack of direct access to the training data distribution as seen by the black-box model.
It can be noticed that by weighing samples according to MS-SSIM, the explanations generated for an image and a distorted version of it are always closer on average than using a cosine distance. The explanations are more robust in the pixel space, in spite of the type of distortion or particular space we sample from.
The same distance values, but normalised to the value of the distances computed by a default LIME configuration, are shown in Fig.~\ref{fig:image-results}. 
It is worth pointing out that if we use a cosine distance, aligning the sampling method to the distortion type of an image results in smaller or equivalent distances than those using the classical LIME approach or other samplings.
Although this is true in the case of the cosine distance, the same does not seem to hold if we use MS-SSIM.


\section{Conclusions and Future Work}
\label{sec:conclusions}

Post-hoc surrogate-based explanation methods like LIME, although widely adopted, rely on sampling from a neighbourhood around a query in order to approximate the local decision boundary learned by a non-interpretable model.
This does not take into account the data distribution with which the black-box model was trained, even though we would ideally sample the neighbourhood to train surrogates from that distribution. In this paper, we showed that for distributions simple enough to be estimated with traditional density estimation techniques, like our 2D example, sampling from the estimated distribution provides a reasonable vicinity to train a surrogate. In order to test the same idea on the visual domain, due to the impossibility in most cases to directly access the true distribution from which the training dataset was sampled (or the distribution of all real images as fall back option), we proposed two complementary approaches to approximate that distribution locally.
On the one hand, we demonstrated that perceptual scale metrics, as they implicitly convey information about the distribution of real images, can be successfully used as a proxy to the distribution of real images, improving the consistency between explanations for similar images.
On the other hand, rather than sampling from patched versions of a query, as is customary, we sampled from distributions of images undergoing some degree of distortion that resemble more closely the set of real images.
We found that in some cases we can further improve the robustness of explanations when combining an aligned sampling method with both euclidean and perceptual distances.

Future research will be devoted to create neighbourhoods not only from interpretable domains based on the visual features of the query, but also on its semantics.
To that end, generative models like DALL-E2~\cite{Ramesh2022HierarchicalTI} or Imagen~\cite{Saharia2022PhotorealisticTD}, both could be used for realistic semantically-driven in-painting of images to better approximate the manifold of real images.




\section{Acknowledgements}
The work leading to these results was supported by the Spanish Ministry of Science and Innovation through the projects GOMINOLA (PID2020-118112RB-C21 and PID2020-118112RB-C22, funded by MCIN/AEI/10.13039/501100011033), and AMIC-PoC (PDC2021-120846-C42, funded by MCIN/AEI/10.13039/501100011033 and by the European Union ``NextGenerationEU/PRTR'').
Ricardo Kleinlein’s research was supported by the Spanish Ministry of Education (FPI grant PRE2018-083225). This work was also funded by UKRI Turing AI Fellowship EP/V024817/1.

\bibliography{egbib}
\end{document}